\documentclass[unnumsec,webpdf,contemporary,large]{oup-authoring-template}%
\usepackage{array, caption, tabularx, makecell, booktabs}%
\usepackage{url,hyperref,lineno,microtype,subcaption}
\usepackage{amsmath} 
\usepackage{amsfonts}
\usepackage{mathtools} 
\usepackage{makecell}
\usepackage{geometry}
\usepackage{amsmath,lipsum}
\usepackage{cuted}%%\stripsep-3pt
\setcellgapes{3pt}

\graphicspath{{Fig/}}

\theoremstyle{thmstyleone}%
\theoremstyle{thmstyletwo}%
\theoremstyle{thmstylethree}%

\begin{document}

\journaltitle{Bioinspiration \& Biomimetics}
\DOI{DOI HERE}
\copyrightyear{2024}
\pubyear{2024}
\access{Advance Access Publication Date: Day Month Year}
\appnotes{Research Paper}

\firstpage{1}

%\subtitle{Subject Section}

\title{Wall-Climbing Performance of Gecko-inspired Robot with Soft Feet and Digits  enhanced by Gravity Compensation}

% \title{Wall Climbing Performance in Gecko-inspired Robot with Gravity Compensation based on Effective Leg-foot Stiffness Estimation}
% \title{Wall-Climbing Stability of Gecko-inspired Robot with Soft Feet by altering effective Leg Stiffness}
% Undulatory Swimming Performance and Body Stiffness Modulation in a Soft Robotic Fish-Inspired Physical Model
% \author[1,2]{Hadrien Sprumont\ORCID{0009-0004-1652-5079}}
% \author[1]{Federico Allione\ORCID{0000-0002-5282-2888}}
% \author[1]{Fabian Schwab\ORCID{0000-0001-9082-8458}}
% \author[1,3]{Bingcheng Wang\ORCID{0000-0002-3258-9812}}
% \author[4]{Claudio Mucingat\ORCID{0000-0003-2676-2815}}
% \author[4]{Ivan Lunati\ORCID{0000-0002-3205-7429}}
% \author[5]{Torsten Scheyer}
% \author[2]{Auke Ijspeert\ORCID{0000-0003-1417-9980}}
% \author[1,3,5,$\ast$]{Ardian Jusufi\ORCID{0000-0003-4250-2984}}
\author[1,3]{Bingcheng Wang}
\author[1]{Zhiyuan Weng}
% \author[1]{Shuangjie Wang}
\author[1]{Haoyu Wang}
\author[1]{Shuangjie Wang}
\author[1,2]{Zhouyi Wang*}
\author[1]{Zhendong Dai}
\author[3]{Ardian Jusufi*}

\authormark{Wang et al.}

% bPaläontologisches Institut und Museum, Universität Zürich, CH-8006 Zürich, Switzerland

\address[1]{\orgdiv{College of Mechanical \& Electrical Engineering}, \orgname{Nanjing University of Aeronautics and Astronautics}, \orgaddress{\postcode{210000}, \state{Jiangsu}, \country{China}}}
\address[2]{\orgname{Shenzhen Research Institute, Nanjing University of Aeronautics and Astronautics}, \orgaddress{\postcode{518063}, \state{Guangdong}, \country{China}}}
\address[3]{\orgdiv{Institute of Neuroinformatics}, \orgname{University of Zuerich and ETH Zuerich}, \orgaddress{ \postcode{8057}, \state{Zuerich}, \country{Switzerland}}}

\corresp[$\ast$]{Corresponding author. \href{wzyxml@nuaa.edu.cn}{wzyxml@nuaa.edu.cn}, \href{ardian.jusufi@uzh.ch}{ardian.jusufi@uzh.ch}}

% \received{Date}{0}{Year}
% \revised{Date}{0}{Year}
% \accepted{Date}{0}{Year}

\abstract{
% When legged robots possess low leg stiffness and are subjected to gravitational loads, their body posture may deviate from the desired configuration, thus affecting the contact angle between the swinging leg's end-effector and the target surface within the gait cycle. This study investigates the relationship between the desired and actual positions of foot-end control in a leg-enhanced stiffness model under external forces to address the challenge of unreliable attachment between the swinging leg's end-effector and the target surface in robots with low leg stiffness. It proposes a feedforward gravity compensation (FGC) strategy with leg coordination, aiming to rectify the body posture influenced by gravity. The strategy seeks to enhance stability during adhesion by reducing body inclination. Finally, a comparative experiment involving a quadrupedal climbing robot EF-\uppercase\expandafter{\romannumeral1} on an inverted surface (ceiling walking), with and without using the FGC strategy, validates the feasibility and effectiveness of the approach. The robot EF-\uppercase\expandafter{\romannumeral1} can operate independently without external assistance. Implementing this strategy addresses the problem of swinging leg end-effectors failing to reliably adhere to the target surface in low-leg-stiffness-legged robots effectively.
Gravitational forces can induce deviations in body posture from desired configurations in multi-legged arboreal robot locomotion with low leg stiffness, affecting the contact angle between the swing leg's end-effector and the climbing surface during the gait cycle. The relationship between desired and actual foot positions is investigated here in a leg-stiffness-enhanced model under external forces, focusing on the challenge of unreliable end-effector attachment on climbing surfaces in such robots. Inspired by the difference in ceiling attachment postures of dead and living geckos, feedforward compensation of the stance phase legs is the key to solving this problem. A feedforward gravity compensation (FGC) strategy, complemented by leg coordination, is proposed to correct gravity-influenced body posture and improve adhesion stability by reducing body inclination. The efficacy of this strategy is validated using a quadrupedal climbing robot, EF-\uppercase\expandafter{\romannumeral1}, as the experimental platform. Experimental validation on an inverted surface (ceiling walking) highlight the benefits of the FGC strategy, demonstrating its role in enhancing stability and ensuring reliable end-effector attachment without external assistance. In the experiment, robots without FGC only completed in 3 out of 10 trials, while robots with FGC achieved a 100\% success rate in the same trials. The speed was substantially greater with FGC, achieved 9.2 mm/s in the trot gait. This underscores the proposed potential of FGC strategy in overcoming the challenges associated with inconsistent end-effector attachment in robots with low leg stiffness, thereby facilitating stable locomotion even at inverted body attitude.
%The same robotic platform would allow researchers to recreate extinct swimming behavior.
}

\newpage

\maketitle

\section{Introduction}

Reliable foot placement and attachment is critical for climbing robots functionality. Essential for these robots are robust surface interaction and the ability to withstand motion-induced disturbances, such as gravitational forces and temporary load that may cause falls or the impacts from attachment-detachment state switching. Various attachment strategies exist, including suction \citep{Guan2013}, magnetic adhesion \citep{Hong2022}, mechanical grips like hooks and claws \citep{parness2017,kim2005spinybotii}, dry adhesion techniques \citep{Hawkes2011}, and wet adhesion methods \citep{wiltsie2012controllably}. Irrespective of the chosen mechanism, legged robots require meticulously designed end-effectors and control systems to ensure dependable adherence across diverse surfaces.

\begin{figure*}
\centering
  \includegraphics[width=0.7\linewidth]{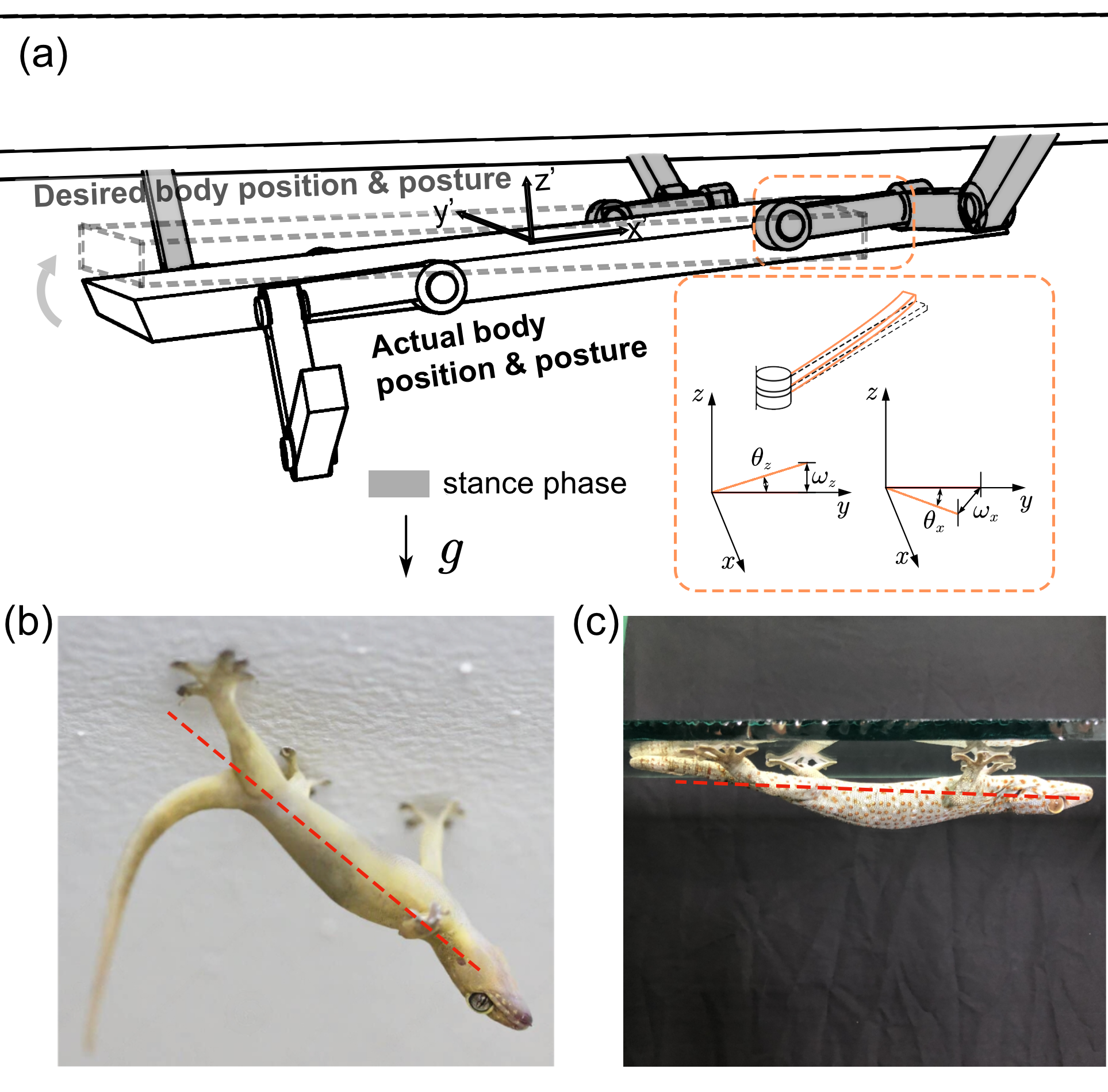}
  \caption{(a) The inclination of the quadruped robot’s body with the ambling gait in a 180-degree plane. The orange section highlights the deformation of the links, this leads to the tilting of the body. At this time, when the swing leg attempts to adhere again, not only can the adhesion position not reach the ideal position, but the posture also deviates. (b) The body posture of a dead gecko on the ceiling.
(c) The body posture of an alive gecko on the ceiling.}
  \label{fig:inc}
\end{figure*}

In the mechanical design of mobile robots, a prevalent trend is to minimize mass by reducing the cross-sectional area of links \citep{klimchik2014stiffness, pashkevich2011enhanced}. The use of lightweight materials has enhanced dynamic performances, particularly in reducing stiffness. The primary effect of decreased stiffness is deformation in the robot's legs or joints under external forces such as gravity or payloads, leading to body tilt. While body tilt is not critical for upright walking robots due to passive gravitational stabilization, it poses a significant challenge for robots climbing surfaces beyond 90 degrees. In these scenarios, as shown in Fig.~\ref{fig:inc} (a), the end-effector of the swing leg may fail to securely attach to the target surface, as observed in ASTERISK \citep{sison2021spherical}, with errors exacerbated by open-loop gait patterns.

To tackle the issue of end-effectors on swing legs failing to reliably attach to climbing surfaces, various methods have been devised to adjust their orientation. This is often achieved through the addition of passive adaptive mechanisms. An underactuated solution involved using a passive spring sufficed for appendage inertia to have a Stickybot outline for body attitude change from rapid inversion from inverted supine position to prone rightside up posture \citep{jusufi2010righting}. Other solutions the end-effector or the incorporation of active actuators, coupled with sensors, to precisely control the end-effector's orientation. For example, the Stickybot III \citep{Hawkes2011} features a passive suspension structure in its ankle joints, which provides misalignment correction. ASTERISK \citep{sison2021spherical}, a robot equipped with magnetic foot suction, incorporates a spherical magnetic joint, enabling angle adjustment for optimal attraction by tilting an adjustable sleeve mechanism. UNIclimb employs a multi-layered structure with a thick sponge layer to better adapt to the contact surface and maximize the adhesive contact area \citep{ko2017wall}. Precise adjustment of the end-effector's posture, as demonstrated by W-Climbot, involves an active wrist joint and an autonomous pose detection module to align suction cups with the target surface, facilitating reliable attachment \citep{guan2012modular}.

Adjusting the posture of an end-effector to align with the target surface is crucial, yet both passive and active methods come with  their distinct benefits and drawbacks. Passive mechanisms can simplify control and require no additional actuators or sensors, thus making them beneficial for preventing unintentional detachments. However, their adaptability is limited, potentially diminishing the maximum adhesion capacity and rendering them ineffective for significant end-effector angle adjustments \citep{Hawkes2011}. Conversely, active adjustments offer precise control over the end-effector's position and orientation, accommodating various target surfaces. Yet, this precision comes at the cost of increased complexity, weight, energy demands, and expenses, making it less viable for smaller robots. 

The sensitivity of robotic stability to preload forces and the positioning of legs relative to the robot's center of mass (CoM) \citep{chen2020inverted}. Their findings indicate that robots are incapable of indefinite inverted surface locomotion without sensory input and feedback mechanisms. In response, they developed an adhesion approach for the robot HAMR, leveraging capillary adhesion and lubrication principles. This design allows for surface sliding while attached, eliminating the need for constant leg lifting and mitigating body tilt issues. However, this adaptation compromises the legged robot's versatility across varied terrains, aligning it more closely with wheeled robot functionalities. 

The quadruped robot HyQ \citep{focchi2017high} manages its weight distribution across its stance legs. Unlike ground-based robots, climbing robots exhibit lower fault tolerance, with gravity's direction posing challenges in maintaining consistent foot-to-surface contact. Consequently, frequent adjustments to body posture through the repositioning of support legs are impractical. Furthermore, feedback control incurs additional demands for sensors and computational resources in the controller, presenting substantial challenges for miniaturized climbing robots. In fact, even if the robot's Inertial Measurement Unit (IMU) can detect the deviation of the body posture, the deviation of the body position still cannot be detected.

In a chance observation, we discovered that although dead geckos can still adhere to the ceiling, their bodies visibly tilts (shown in Fig.~\ref{fig:inc} (b)), similar to the robot in Fig.~\ref{fig:inc} (a). This phenomenon has never been observed in living geckos (shown in Fig.~\ref{fig:inc} (c)). We believe that the tilting of the body in the former is due to the loss of stimulation in the muscles of the stance phase legs, which results in decreased joint stiffness, whereas living geckos can actively adjust the state of their stance phase legs, thus keeping themselves close to the surface. Inspired by this, we propose a novel feedforward gravity compensation algorithm. By establishing the leg-enhanced stiffness model, we calculate the stance phase legs end position after gravity compensation. This study focuses on a legged climbing robot and addresses the issue of body inclination caused by the gravitational load during the gait cycle. Specifically, the study investigates the impact of the end-effector's contact pose with the target surface and its effect on reliable attachment.

Firstly, we analyzed the deformation of the legs, established the leg-enhanced stiffness model, and then analyzed the relationship between the leg configuration, the stance phase under external force, and the body’s posture. Then, use quadratic programming to optimize the foot force, derive the relationship between the foot control position and the actual position under external force, and propose the feedforward gravity compensation strategy with leg coordination. Foot force is the force exerted on the distal end of the last link in the robot leg model, and its coordinate system direction is the same as the body coordinate system. Finally, an inverted surface ambling gait crawling experiment was conducted on the quadruped climbing robot platform, and the strategy’s feasibility was verified through data.

\section{Control Strategy}

In legged robot static mechanical analysis, joints and linkages are typically considered as rigid. However, for miniature robots, the linkage deformation and joint stiffness should be considered, which cause significant posture deviations under gravitational loads and resulting in body tilting. This is particularly noticeable when a multi-legged robot navigates large inclined planes, as illustrated for a quadruped wall-climbing robot using an ambling gait in Fig.~\ref{fig:inc}. Such deviations impact the end-effector's contact angle with the target surface, a critical factor for climbing robots that require precise end-effector positioning, making leg deformations a non-negligible factor.

This section presents a static mechanical analysis and the generation of compensation items, focusing on enhanced leg stiffness, and introduces a gravity compensation-based control strategy for correcting body posture.

\subsection{Extended Leg Stiffness Model}

To facilitate the analysis, several assumptions regarding the linkages are made: 1) The material composition of the linkages is homogeneous throughout. 2) Bending deformations in the linkages significantly surpass those from torsion and compression. 3) Buckling phenomena are absent in the linkages.

From assumption (1), it is deduced that stress is uniformly distributed across the linkages. In a state of static equilibrium, the leg linkage is considered fixed at one end and free at the other. Forces applied to the free end, coupled with assumption (2), justify treating the linkage as a cantilever beam. It is important to note that when deformation occurs, the end coordinate system undergoes not just translation but also rotation in response to the endpoint's movement.

Deflection, denoted as $\omega$, is the linear displacement of the linkage in a direction perpendicular to its axis. Assumption (3) guarantees that deflection undergoes no sudden changes. Section rotation, $\theta$, refers to the rotation angle of the member's cross-section from its original orientation. Based on material mechanics principles, the equation $\omega = -\frac{fl^3}{3EI}$ is derived, applicable when force $f$ acts perpendicular to the linkage's direction.

To address the fuselage tilt issue, the deformation of the leg linkage is identified as the primary cause. Consequently, variables representing leg bending, such as sectional rotation angles $\boldsymbol{\theta}_i$ and deflection $\boldsymbol{\omega}i$, are incorporated. These variables introduce an additional rotation-translation relationship between each joint, expressed as $\mathbf{T}_{si}(\boldsymbol{\theta}_i,\boldsymbol{\omega}_i)$. Here, $\boldsymbol{\theta}_i=(\theta_x,\theta_z)^\top$ denotes the sectional rotation angles due to forces along the x-axis and z-axis, respectively, and $\boldsymbol{\omega}_i=(\omega_x,\omega_z)^\top$ represents the deflection from forces along the same axes. The linkage's coordinate system is illustrated in Fig.~\ref{fig:inc}, with $i$ indicating the joint number.

Utilizing the revised Denavit-Hartenberg (DH) parameters \citep{Denavit1955}, the homogeneous transformation matrix from connected coordinate system $i-1$ to coordinate system $i$ is represented by $\prescript{i-1}{i}{\boldsymbol{A}}$. The rotation-translation transformation matrix between each joint, including $\mathbf{T}_{si}$, is denoted as $\prescript{i-1}{i}{\boldsymbol{T}}$. By accounting for the passive joint rotation at the end effector, the final pose transformation from the end-effector coordinate system to the shoulder joint coordinate system is derived.

\begin{equation}\boldsymbol{T}= \prescript{0}{n}{\boldsymbol{T}}\boldsymbol{R}_s(\boldsymbol{q}_s)
\label{eq:1}
\end{equation}

where $\boldsymbol{R}s(\cdot)$ indicates the rotation matrix for the end effector's passive joints, $\boldsymbol{q}s=(\varphi{sx}, \varphi{sy}, \varphi_{sz})^\top$ denotes the passive joint rotation variables, and $n$ represents the total number of joints. The detailed expansion of equation (\ref{eq:1}) is provided in the supplement document 1.

By applying equation (1), the end-effector pose $\boldsymbol{\varGamma} = \boldsymbol{g}(\boldsymbol{\omega}, \boldsymbol{q}, \boldsymbol{q}s)$ in the shoulder coordinate system is determined by the joint variables $\boldsymbol{\omega} = (\omega{1x}, \omega_{1z}, \omega_{2x}, \omega_{2z}, ..., \omega_{nx}, \omega_{nz})^\top$, $\boldsymbol{q} = (q_1, q_2, ..., q_n)^\top$, and $\boldsymbol{q}s = (\varphi_{sx}, \varphi_{sy}, \varphi_{sz})^\top$, where the end-effector pose is represented as $\boldsymbol{\varGamma} = (x, y, z, \varphi_x, \varphi_y, \varphi_z)^\top$.

\subsection{Static Analysis}
For climbing robots, it is reasonable to assume that, aside from moments of leg motion phase transitions, the robot maintains a state of static equilibrium. Based on this premise, a static analysis of the robot will be undertaken.

According to the principle of virtual work, in a state of static equilibrium, the total virtual work done by all forces for any virtual displacement equals zero. Thus, the equilibrium condition is established as:

\begin{equation}
    \left[ \begin{array}{c}	\boldsymbol{f}_{\omega}\\	\boldsymbol{\tau }_q\\	\boldsymbol{\tau }_{qs}\\\end{array} \right] =\left[ \begin{array}{c}	\boldsymbol{J}_{\omega}^{\top}\\	\boldsymbol{J}_{q}^{\top}\\	\boldsymbol{J}_{qs}^{\top}\\\end{array} \right] \mathbf{F}=\left[ \begin{array}{c}	-\boldsymbol{K}_{\omega}\varDelta \boldsymbol{\omega }\\	-\boldsymbol{K}_q\varDelta \boldsymbol{q}\\	-\boldsymbol{K}_{qs}\varDelta \boldsymbol{q}_s\\\end{array} \right] ,
\end{equation}
where $\boldsymbol{J}_{\omega}\in \mathbb{R} ^{6\times 2n},\boldsymbol{J}_q\in \mathbb{R} ^{6\times n},\boldsymbol{J}_{qs}\in \mathbb{R} ^{6\times 3}$are the kinematic Jacobian matrices with respect to the variables $\boldsymbol{\omega }$, $\boldsymbol{q}$ and $\boldsymbol{q}_s$ respectively. The external forces and torques $\mathbf{F}=\left( f_x,f_y,f_z,m_x,m_y,m_z \right) ^{\top}$ applied to the end-effector. The internal forces include forces $\boldsymbol{f}_{\omega}\in \mathbb{R} ^{2n}$ due to linkage deformations, torques $\boldsymbol{\tau }_q\in \mathbb{R} ^n$ generated by active rotational joint deformations, and torques $\boldsymbol{\tau }_{qs}\in \mathbb{R} ^3$ generated by passive joint deformations. And $\varDelta \boldsymbol{\omega }, \varDelta \boldsymbol{q}, \varDelta \boldsymbol{q}_s$ represent the increment of the variables after the application of forces, and the relationship between force and deflection can be determined using $\omega =-\frac{fl^3}{3\mathrm{EI}}$, which is $\boldsymbol{K}_{\omega}=\mathrm{diag}\left( \frac{3EI_{1x}}{{L_1}^3},\frac{3EI_{1z}}{{L_1}^3},...,\frac{3EI_{ix}}{{L_i}^3},\frac{3EI_{iz}}{{L_i}^3} \right) $.

\subsection{Gravity Compensation with Legs Coordination}

When the end-effectors of a climbing robot maintain non-slipping contact with the climbing surface, the robot's body posture relative to the climbing plane is dictated by the positions of the stance feet in relation to the robot's body. Thus, correcting the body posture is essentially about compensating for the stance feet's positions. Consequently, the desired body posture aligns with the positions of the end-effectors during the stance phase in relation to the shoulder joints of the robot, establishing known target foot positions.

By conducting a mechanical analysis of the robot in the desired body posture, the force and torque equilibrium equations for the robot can be derived.

\begin{equation}
	\begin{cases}	\sum_{i=1}^n{\boldsymbol{S}_i}\boldsymbol{f}_{ci}+\boldsymbol{F}_{CoM}=\mathbf{0}\\	\sum_{i=1}^n{(}\boldsymbol{S}_i\boldsymbol{p}_{fci}\times \boldsymbol{f}_{ci})=\mathbf{0}\\\end{cases}   ,
 \label{eq:3}
\end{equation}
where $\boldsymbol{f}_{ci}\in \mathbb{R} ^3$ represents the force exerted on the $i$-th leg's end-effector, $\boldsymbol{F}_{CoM}\in \mathbb{R} ^3$ denotes the external force acting on the robot's center of mass (CoM), $\boldsymbol{S}_i\in \mathbb{R} ^{3\times 3}$ is the selection matrix. When the $i$-th leg is in the stance phase, $\boldsymbol{S}_i=\mathbf{I}$. When the $i$-th leg is in the swing phase, $\boldsymbol{S}_i=\mathbf{0}$ . And $n$ represents the number of legs of the robot, $\boldsymbol{p}_{fci}\in \mathbb{R} ^3$ is the spatial vector from the $i$-th leg's end-effector to the robot's COM.

This system consists of 6 equations and $k=3m$ unknown variables, where $m$  represents the least number of legs in the stance phase. Therefore, when $m=2$, the equation has a unique solution, and when $m>2$ , the system of equations has an infinite number of solutions. Hence, it is essential to seek the optimal solution based on the constraints imposed by the forces at the end effectors when $m>2$.

The transformation of two vector cross-products into matrix and vector dot products is possible. For example, $\boldsymbol{p}_{fci}\times \boldsymbol{f}_{ci}$ can be represented as $\left[ \boldsymbol{p}_{fci}\times \right] \boldsymbol{f}_{ci}$ , which is the dot product of the left cross-product matrix $\left[ \boldsymbol{p}_{fci}\times \right] $ and vector $\boldsymbol{f}_{ci}$ . Consequently, equation (\ref{eq:3}) can be reformulated as

\setlength{\arraycolsep}{1pt}
\begin{equation}
      \mathop {\underbrace{\left[ \begin{matrix}	\mathbf{I}_3&		...&		\mathbf{I}_3\\	\left[ \boldsymbol{p}_{fc1}\times \right]&		...&		\left[ \boldsymbol{p}_{fcn}\times \right]\\\end{matrix} \right] }} \limits_{\boldsymbol{A}}\underset{\boldsymbol{f}}{\underbrace{\left[ \begin{array}{ccc}	\boldsymbol{S}_1\boldsymbol{f}_1 &		\mathbf{0} &		\mathbf{0}\\	\mathbf{0} &		...&		\mathbf{0}\\	\mathbf{0}&		\mathbf{0}&		\boldsymbol{S}_n\boldsymbol{f}_n\\\end{array} \right]  }}=-\underset{\boldsymbol{F}_c}{\underbrace{\left[ \begin{array}{c}	\boldsymbol{F}_{CoM}\\	\mathbf{0}\\\end{array} \right] }}. 
      \label{eq:4}
\end{equation}

 Based on the constraints outlined in the above equation, it is possible to formulate an objective function. By transforming the objective function into the standard form of a quadratic function (\ref{eq:5}), quadratic programming can be employed to compute forces on each end-effector in the stance phase to achieve coordination of the legs in the stance phase.
\begin{equation}
    \begin{aligned}	
    \boldsymbol{f}^d=&\underset{\boldsymbol{f}\in \mathbb{R} ^k}{\mathrm{arg}\min}\left\{ (\boldsymbol{Af}+\boldsymbol{F}_c)^{\top}\boldsymbol{S}(\boldsymbol{Af}+\boldsymbol{F}_c)+\alpha \boldsymbol{f}^{\top}\boldsymbol{Wf} \right\}\\	&\,\,\mathrm{s}.\mathrm{t}.\quad \underline{\boldsymbol{d}}<\boldsymbol{f}<\bar{\boldsymbol{d}}\\
    &\Rightarrow \begin{cases}	\boldsymbol{f}^d=\underset{f\in \mathbb{R} ^n}{\mathrm{arg}\min}\left\{ \frac{1}{2}\boldsymbol{f}^{\top}\boldsymbol{Hf}+\boldsymbol{f}^{\top}\boldsymbol{g} \right\}\\	\boldsymbol{H}=2\left( \boldsymbol{A}^{\top}\boldsymbol{SA}+\alpha \boldsymbol{W} \right)\\	\boldsymbol{g}=2\boldsymbol{A}^{\top}\boldsymbol{SF}_c\\\end{cases}
    \end{aligned}
    \label{eq:5}
\end{equation}
\noindent where $\boldsymbol{f}^{\top}\boldsymbol{Wf}$ is a cost function for evaluating the summation of end-effector forces. Additional evaluation can include the peak end-effector force, the distance from the CoM to the plane, and more. And $\boldsymbol{f}^d\in \mathbb{R} ^3$ represents the end-effector forces when the objective function takes the minimum value. $\boldsymbol{A}\in \mathbb{R} ^{6\times 3n}$ is the matrix associated with the force and torque equilibrium equations, $\boldsymbol{f}\in \mathbb{R} ^{3n}$ denotes all end-effector forces, $\boldsymbol{F}_c=\left[ \begin{matrix}	\boldsymbol{F}_{CoM}&		\mathbf{0}\\\end{matrix} \right] ^{\top}\in \mathbb{R} ^6$ including external forces acting on the CoM, $\boldsymbol{S}\in \mathbb{R} ^{6\times 6}$ and $\boldsymbol{W}\in \mathbb{R} ^{n\times n}$ are weight matrices which are positive definite, $\boldsymbol{S}$ is used to adjust whether the robot focuses more on force distribution or force/torque balance. The larger the $\boldsymbol{S}$, the more important the force/torque balance becomes, but the effectiveness of the robot's force distribution will worsen. $\underline{\boldsymbol{d}}, \bar{\boldsymbol{d}}\in \mathbb{R} ^n$ representing the lower and upper bounds for the end-effector forces, respectively, and $\alpha \in \mathbb{R} $ is the quadratic regularization coefficient. Notably, adjusting the coefficient $\alpha $ during quadratic programming is a critical step, as excessive regularization can lead to significant tracking errors, thus adversely affecting the robot's balance.

The initial angle of the actively rotating joint without considering leg deformation is represented as $\boldsymbol{q}_0$ . The target position $\boldsymbol{q}_t$ corresponds to the end-effector position after accounting for the effects of leg stiffness. When the forces acting on the end-effector in the stance phase are known, and the target positions of the end-effector are known calculating the deformation parameters and joint angles of one leg is possible. This allows for the determination of command variables. By combining (\ref{eq:4}) and the target foot position formula, the following equation can be obtained.

\begin{equation}
    	\left\{ \begin{array}{c}	\left[ \begin{array}{c}	\boldsymbol{K}_{\omega}\boldsymbol{\omega }\\	\boldsymbol{K}_q\left( \boldsymbol{q}_t-\boldsymbol{q}_0 \right)\\	\boldsymbol{K}_{qs}\boldsymbol{q}_s\\\end{array} \right] +\left[ \begin{array}{c}	\boldsymbol{J}_{\omega}^{\top}\\	\boldsymbol{J}_{q}^{\top}\\	\boldsymbol{J}_{qs}^{\top}\\\end{array} \right] \mathbf{F}=0\\	\boldsymbol{\varGamma }_t=\boldsymbol{g}\left( \boldsymbol{\omega }_t,\boldsymbol{q}_t,\boldsymbol{q}_s \right)\\\end{array} \right.      
     \label{eq:6}
\end{equation}

When the target end-effector pose $\boldsymbol{\varGamma }_t$ is known, the stiffness coefficients $\boldsymbol{K}_{\omega}, \boldsymbol{K}_q, \boldsymbol{K}_{qs}$ are known, and the forces and torques acting on the end-effector $\boldsymbol{F}$ are known, the value of $\boldsymbol{\omega }_t, \boldsymbol{q}_t, \boldsymbol{q}_0, \boldsymbol{q}_s$ can be obtained by (\ref{eq:6}).

\begin{figure*}[htbp]
    \centering
    \includegraphics[width=0.8\linewidth]{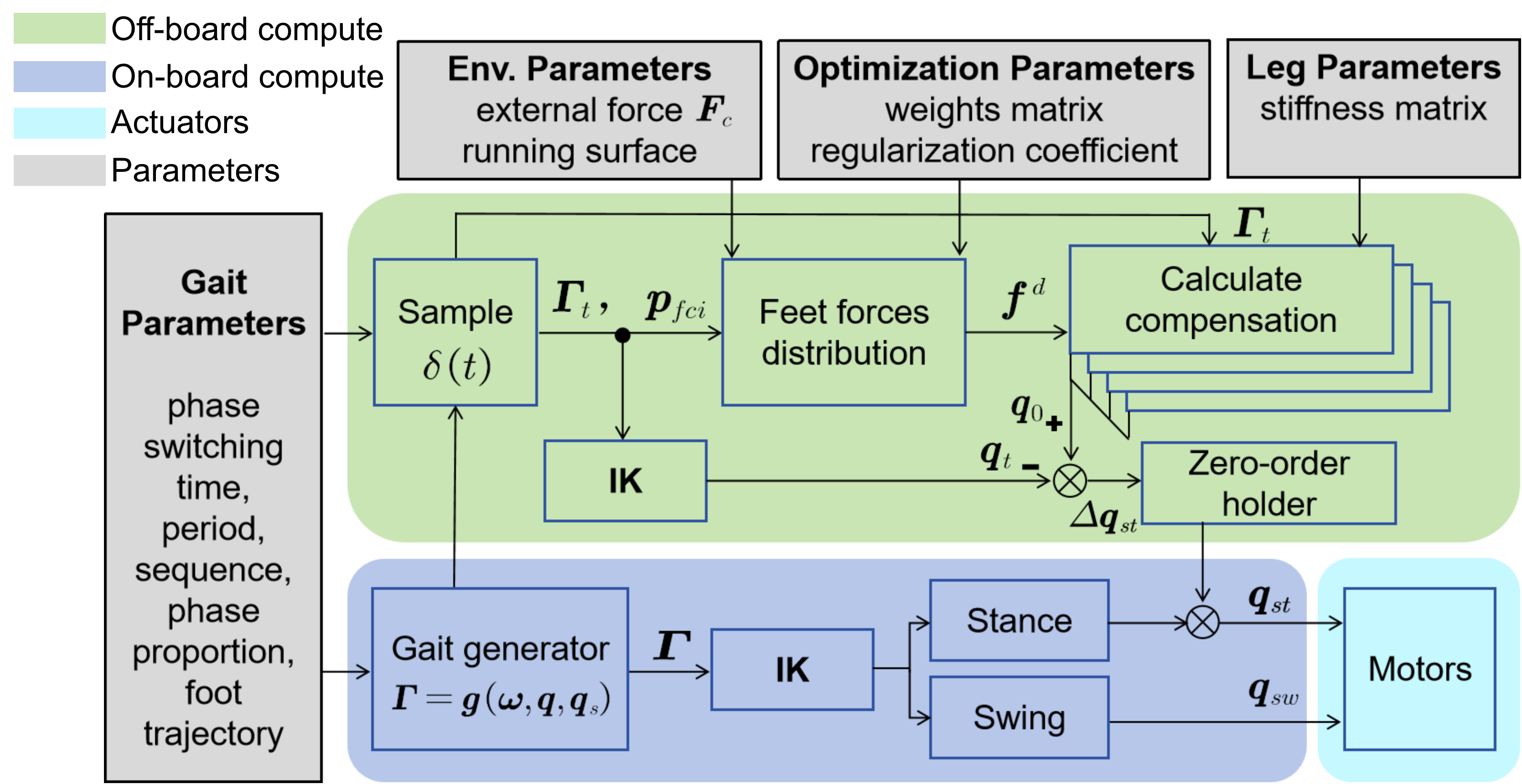}
    \caption{Overall control framework. The green section represents off-board computation, after the calculation is completed, the offset can be imported into the on-board computation as the preset. The dark blue section represents feedforward compensation for real-time inverse solutions of the gait generator. The gray section represents the parameters input layers, These parameters need to be measured in advance or artificially set according to the motion requirements. The light blue section represents the motors.}
    \label{fig:diagram}
\end{figure*}

The Feedforward Gravity Compensation (FGC) strategy is proposed to calculate offline and compensate in advance, as shown in Fig.~\ref{fig:diagram}.  The essence of FGC strategy is to achieve the target foot position by calculating the joint compensation of the deformed support phase leg. In blue part of Fig.~\ref{fig:diagram}, gait generator generates target foot position for each leg based on gait parameters. The part ‘IK’(inverse kinematics) obtains joint control angles by inverse kinematics solution of target foot positions. If the joint control angles are directly input to the servo system, this is a basic position control method. The above process is one step in the operation of the robot, and the gait generator continuously generates the target foot positions to make the robot continue move. The green section of Fig 2 is the key of FGC strategy. To reduce computational burden, the gait generator is sampled and the compensation for joint angles is calculated offline. Perform force distribution with sampled target foot positions $\boldsymbol{\varGamma }_t$ by (5) quadratic program based on environmental and optimization parameters, the target foot forces $\boldsymbol{f}^d$ can be obtained. Then we can use the equation (6) to calculate the corrected joint control angles $\boldsymbol{q}_0$ of the extended leg stiffness model. The sampling frequency is different from the running frequency of the gait generator, so the joint control angle cannot be directly input into the servo system. It requires calculating the joint angle compensation $\varDelta \boldsymbol{q}_{st}$ and connecting it to the output of the gait generator through a zero order holder. The joint angle compensation is obtained by $\varDelta \boldsymbol{q}_{st}=\boldsymbol{q}_0-\boldsymbol{q}_t$ and the original joint control angle $\boldsymbol{q}_t$ is obtained by sampling the target foot position through part ‘IK’.

In the offline computation part, all parameters are known, and the input is obtained through sampling by the gait generator, e.g., the positions of end-effectors in stance phase for the desired robot body pose. The sampling frequency should not be too low, ensuring at least one sampling point between each phase. Once the input for offline computation is obtained, end-effector force distribution is carried out based on environmental parameters and optimization parameters. Subsequently, the motor command angles $\boldsymbol{q}_0$ are computed using a leg-enhanced stiffness model, based on the end-effector forces and leg parameters. Finally, the compensation parameter for the stance phase $\varDelta \boldsymbol{q}_{st}$ is obtained at the sampling moment. This compensation parameter is added to the motor angles output by the gait generator for the stance phase, along with the motor angles for the swing phase, and then sent to the motors to achieve posture correction.

In order to compare the effectiveness of FGC strategy, a basic control method without using FGC strategy was designed, which only uses blue parts. The gait generator continuously generates the target foot positions $\boldsymbol{\varGamma }$, and joint control angles obtained through ‘IK’ are directly input to the servo system.

\section{Experimental Setup}
\subsection{Quadrupeded Wall-climbing Robot}
\begin{figure*}[htbp]
    \centering
    \includegraphics[width=0.9\linewidth]{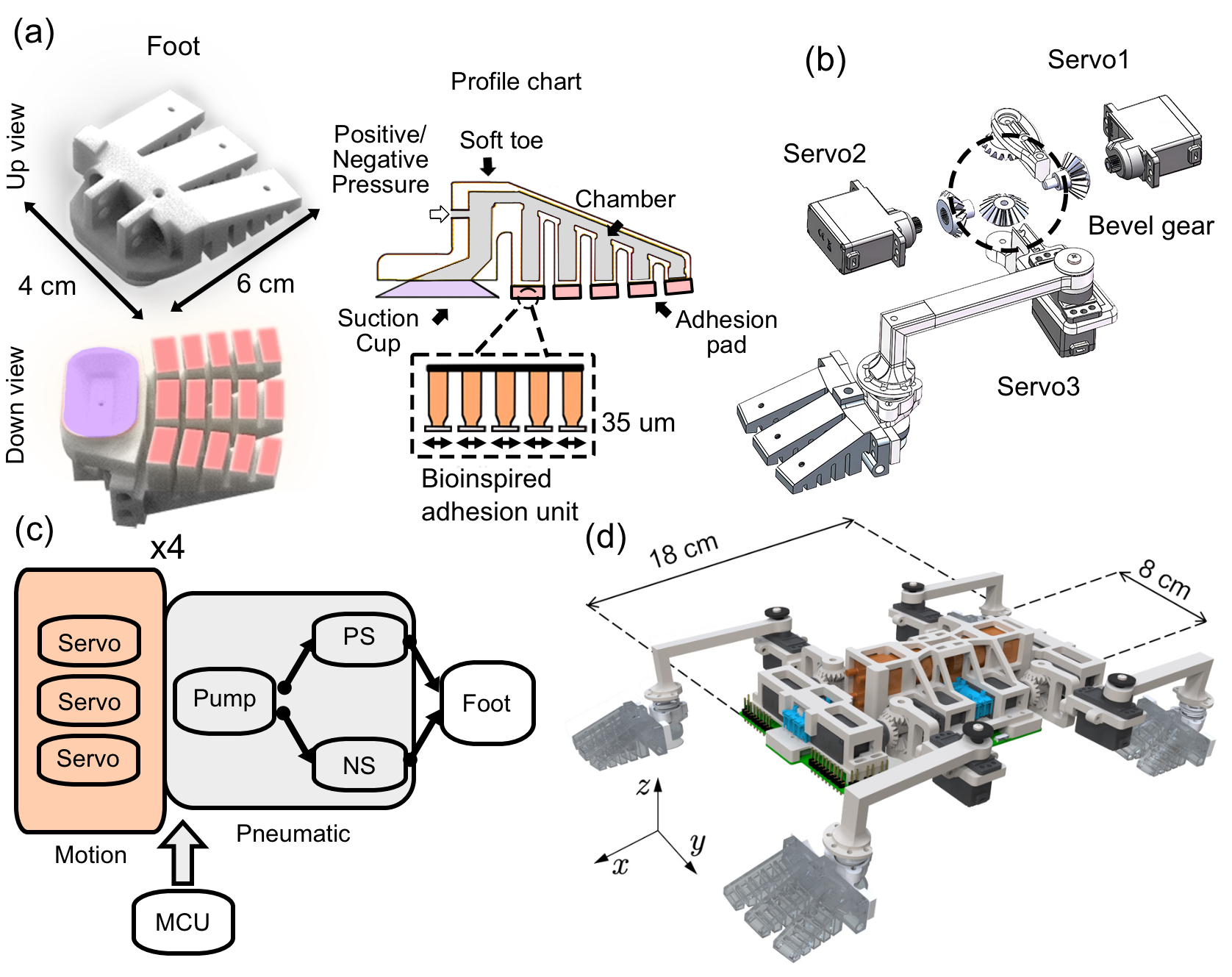}
    \caption{(a) The soft pneumatic foot. When the sole is under negative pressure, the suction cup will generate adhesive force. In addition, the toes will contract to help the adhesion pad attach. Conversely, when the sole is under positive pressure, both the pneumatic and adhesive systems of the sole will lose their ability to adhere. (b) The exploded diagram of the robot's left front leg. The horizontal bevel gear ensures coaxiality through a positioning axis, while the vertical bevel gear is connected to the robot leg, which is connected to a motor outside the body. (c) Schematic diagram of the robot hardware control system. The Micro Controller Unit (MCU) calculates the angle of the servo according to the control flow in Fig.~\ref{fig:diagram}, and controls the on-state of the positive solenoid valve (PS) and the negative solenoid valve (NS) in the corresponding pneumatic system based on the current motion state of the leg. (d) Quadruped climbing robot EF-\uppercase\expandafter{\romannumeral1}. The coordinate system in the Fig is the body coordinate system. }
    \label{fig:rbt}
\end{figure*}

The robot platform used for validating the proposed strategy in this study is the quadruped climbing robot EF-\uppercase\expandafter{\romannumeral1}, powered by electricity and pneumatic, the robot consists of four parts, namely soft  pneumatic foot, 3-degrees-of-freedom bevel-gear-based motion system, control system (respectively shown in Fig.~\ref{fig:rbt} (a, b, c)). This robot features a central symmetric distribution of its four legs and body structure, as shown in Fig.~\ref{fig:rbt} (d). It employs pneumatic coupled adhesive footpads as its end-effectors, which have suction cups and adhesive arrays on the undersides. The adhesive arrays are made of polyvinyl siloxane (PVS) \citep{Gorb2006}, which is similar to gecko bristles, thus, the footpads can stably adhere to a smooth surface. The robot is equipped with a total of 12 motors. Each leg consists of three active joints. The two degrees of freedom of the shoulder are provided by a bevel gear set, namely the lateral swing shoulder joint and the upper lift shoulder joint. The structure of the bevel gear set can place the motors on the body, thereby concentrating the mass on the body. The ankle joint is connected to the end-effector through a spring and ball joint, offering three passive rotational degrees of freedom and enabling limited rebound. 

The leg links are constructed with a hollow structure to make the robot as lightweight as possible. The cross-section of the hollow part and cross-sectional parts of the leg are both rectangular, and the centers of the two cross-sections coincide. The 3D printing material used for the robot is Tough-2000, considering the stiffness, hardness, and density of the material.

Of the 12 motors on the robot, eight are located within the body of the robot, while the remaining four are positioned at the first leg link of each leg. The robot's pump, solenoid valves, battery, and controller are all situated within the body, following a central symmetric distribution. Consequently, most of the robot's mass is concentrated in its body, and it is easy to assume that the robot's CoM coincides with the geometric center of the body. The total mass including battery is 487 g. Despite being miniaturized, lightweight, and carrying all the weight, the robot EF-\uppercase\expandafter{\romannumeral1} can still operate independently without external power supply or weight reduction.

The robot’s end-effector employs pneumatic actuators to achieve attachment by rolling down and gripping the toes inward and detachment by rolling the toes upward and backward \cite{tian2006adhesion, wang2010morphology}. This end-effector inspired by geckos could make attachment and detachment easy, owing to its adhesion-pneumatic coupled mechanism. This flexible foot provides excellent large deformation, making the end-effector more applicable, but reducing leg stiffness. Each footpad can exert a maximum normal end-effector force of up to 8 N and a maximum tangential end-effector force of 3.5 N. This needs the contact area between the foot and the target plane to be as large as possible, which means the plane of the footpad should be as parallel as possible to the target plane.

The robot's ankle joints are designed with ball joints to accommodate the contact between the footpad plane and the climbing surface. Additionally, tiny springs are incorporated at the ball joint to provide retraction, representing the only passive joints in the robot. Due to the low stiffness of these tiny springs, they generate minimal torque during the robot's operation. Therefore, the impact of torque generated by the deformation of these passive joints on the equilibrium equations is negligible. As a result, it is possible to simplify equation (\ref{eq:6}) as follows:

\begin{equation}
\begin{aligned}
    \left[ \begin{array}{c}	\boldsymbol{K}_{\omega}\varDelta \boldsymbol{\omega }\\	\boldsymbol{K}_q\varDelta \boldsymbol{q}\\\end{array} \right] +\left[ \begin{array}{c}	\boldsymbol{J}_{\omega f}^{\top}\\	\boldsymbol{J}_{qf}^{\top}\\\end{array} \right] \mathbf{F}_f=0\\                           	
	\boldsymbol{p}_t=\boldsymbol{g}_p\left( \boldsymbol{\omega }_t,\boldsymbol{q}_t \right) 	
\end{aligned} \label{eq:7}      
\end{equation}
where $\mathbf{F}_f\in \mathbb{R} ^3$ represents the force acting on the end-effector, which is in the stance phase, $\boldsymbol{J}_{\omega f}\in \mathbb{R} ^{3\times 3}$ and $\boldsymbol{J}_{qf}\in \mathbb{R} ^{3\times 3}$ are the Jacobian matrices corresponding to the end-effector position $\boldsymbol{g}_p\left( \boldsymbol{\omega }_t,\boldsymbol{q}_t \right) $ corresponding variables $\boldsymbol{\omega }_t$ and $\boldsymbol{q}_t$, respectively, and $\boldsymbol{p}_t\in \mathbb{R} ^3$ represents the target end-effector position.

Because the bevel gear mechanism at the shoulder/hip joint is driven by two motors simultaneously, the rotational stiffness of these two active rotational joints equals the sum of the rotational stiffness of the two motors. At the elbow/knee joints, each active revolute joint is driven by a single motor, and the joint stiffness is equivalent to the rotational stiffness of the motor at that location. The stiffness of the robot's leg links can be calculated from the relationship between force and deflection or obtained through calibration experiments. Therefore, the stiffness matrix $\boldsymbol{K}_{\omega}$, $\boldsymbol{K}_q$ can be determined.

\begin{figure}
    \centering
    \includegraphics[width=0.8\linewidth]{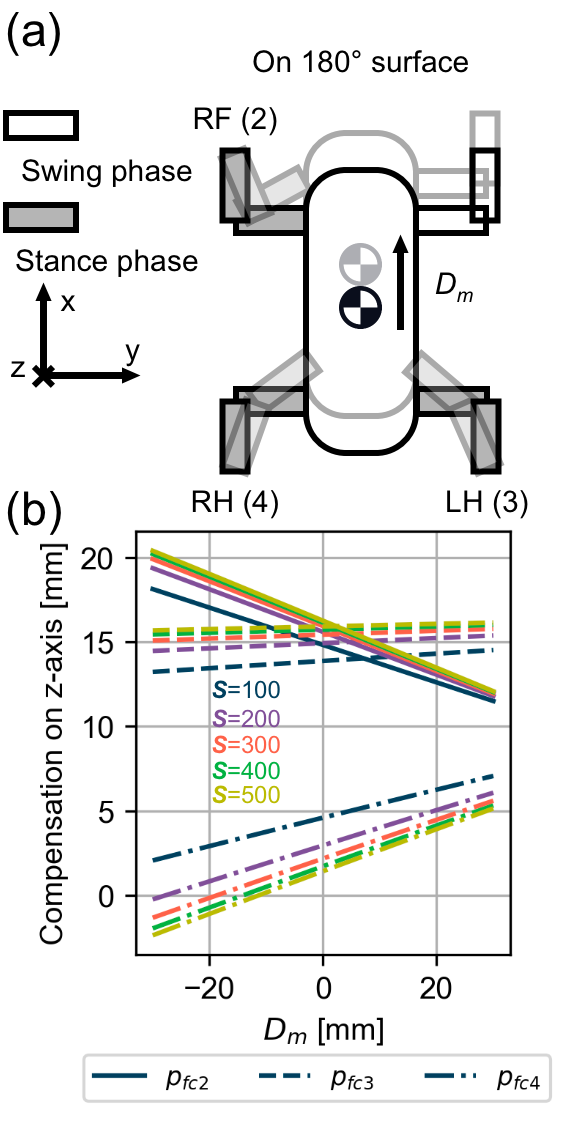}
    \caption{(a) A schematic diagram of 180$^\circ$ surface climbing. It should be noted that due to the 180$^\circ$ surface, the robot's left front leg is shown as the right front leg in the picture. $D_m$ represents the x-axis position of the current location relative to the midpoint of the stance diagonal. (b) The diagram shows the z-axis distance that needs to be compensated by the stance legs under different $D_m$ and $\boldsymbol{S}$ in equation~\ref{eq:5}.}
    \label{fig:comp}
\end{figure}

For a EF-\uppercase\expandafter{\romannumeral1} application, take the example when the left foreleg enters the swing phase at 180 degrees, as shown in Fig.~\ref{fig:comp} (a). During the movement of the robot, the desired position of the support phase relative to the center of mass continuously changes. We have written a MATLAB script for outputting foot position compensation, which has yielded the displacement compensation required in a 180-degree climbing scenario. The compensation is mainly concentrated on the Z-axis, and the results can be seen in Fig.~\ref{fig:comp} (b).

Taking the left front leg of the robot as an example, in its joint coordinate system, the leg deformation variables are denoted as $\omega _{1x}, \omega _{1z}, \omega _{2x}, \omega _{2z}$ , which corresponds to the section rotation angles $\theta _{1x},\theta _{1z},\theta _{2x},\theta _{2z}$. The parameters of the left front leg constructed by the modified D-H method are listed in the supplement document 2.

Based on D-H parameters, the transformation matrix can be calculated, subsequently allowing for determining the end-effector position in the shoulder joint coordinate system. Here, $q_1, q_2, q_3$ represent the variables for the active revolute joint of that leg. By following the approach used for the left front leg, it is possible to establish the entire robot's leg-enhanced stiffness model, thereby obtaining both forward and inverse kinematics solutions for the robot's motion.

\subsection{Experimental Preparation}

\begin{figure}
    \centering
    \includegraphics[width=\linewidth]{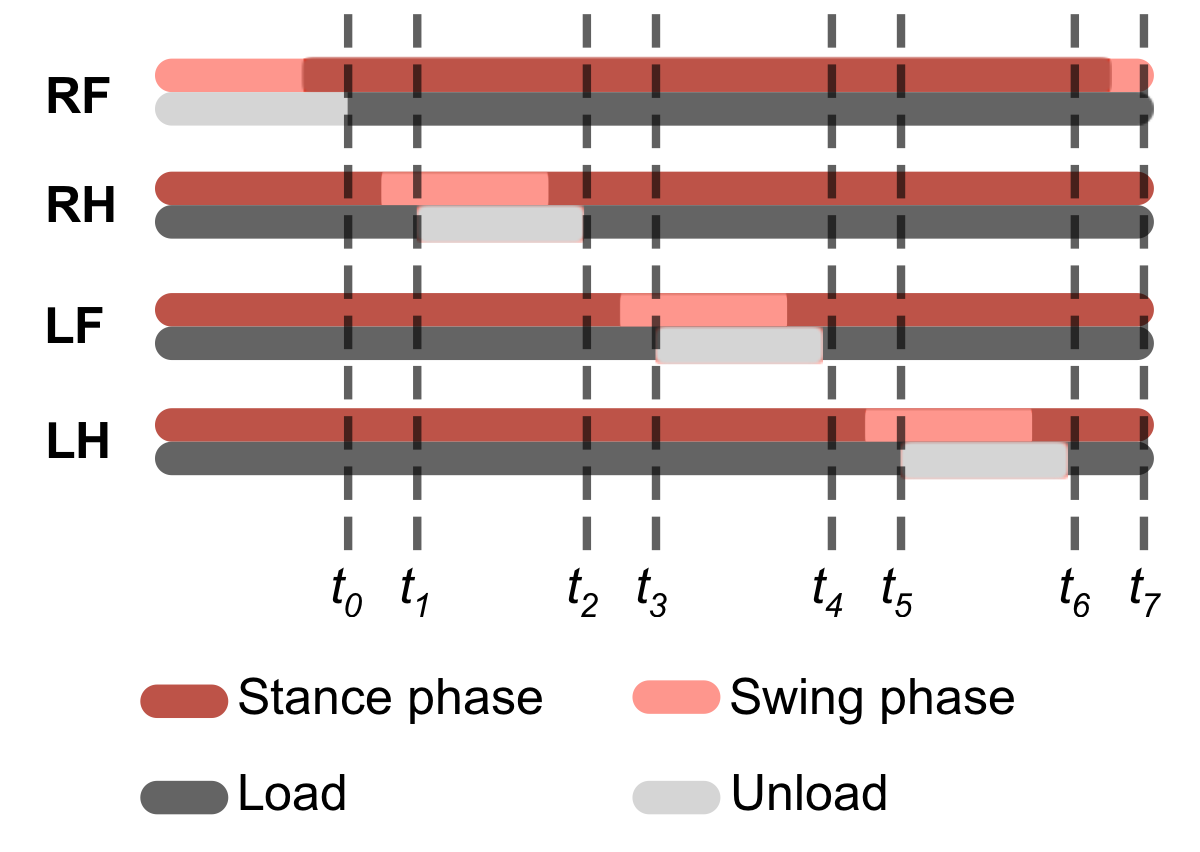}
    \caption{The temporal diagram for the robot's gait planning. Sampling times are selected to coincide with the phase transition between the swing and stance phases, as indicated by the dashed lines. For the robot's entire motion cycle of duration \(T\), the time points are defined as follows: $t_0 = \frac{3}{16}T$, $t_1 = t_0 + \frac{1}{16}T$,$ t_i = t_{i-2} + \frac{1}{4}T (i \in \{2, 3, ..., 7\})$. "load" denotes loading force with the attachment action of the footpad through pneumatic mechanisms, while "unload" denotes unloading force with the detachment action of the footpad through pneumatic mechanisms.}
    \label{fig:gait}
\end{figure}

Trot gait and ambling gait are both classic gaits of quadrupeds \citep{focchi2017high, kim2019highly}. A common trait of the ambling gaits is that usually only one foot is completely off the ground at any one time, one of the timing planning of ambling gait is shown in Fig.~\ref{fig:gait}, and in trot gait both legs on the diagonal are in the swing or stance phase simultaneously. Based on the calculation results of equation (\ref{eq:5}), it is not difficult to find that in the trot gait, the position of the robot will shift, and in the ambling gait, both the posture and position will shift. Therefore, the ambling gait is used in this section to validate the FGC strategy proposed in the previous subsection. Sampling times are selected to coincide with the phase transition between the swing and stance phases. It should be noted that, our algorithm is also applicable in the trot gait, and the motion video of the robot's trot gait can be seen in the supplement video 2.

Given that the end-effector of the robot used in this study is equipped with a suction cup at its bottom center, the condition for ensuring a reliable attachment of the end-effector to the climbing surface is to align the plane of the end-effector with the climbing plane. Because the robot's legs are identical and the shoulder joints are in the same plane as the body, and since the climbing surface is planar, the legs in the stance phase are in the same plane and parallel to the climbing surface. As a result, the desired body posture for the robot in each phase is for the body to be parallel to the climbing plane.

In the experiments, the robot moves uniformly forward along the positive direction of the x-axis of the body coordinate system. We have obtained the gait parameters of the robot, including phase transition times, leg lifting sequence, phase ratios, and foot trajectories, which could influence the stability of robot crawling \citep{wang2023neural}. 

\begin{figure*}
    \centering
    \includegraphics[width=\linewidth]{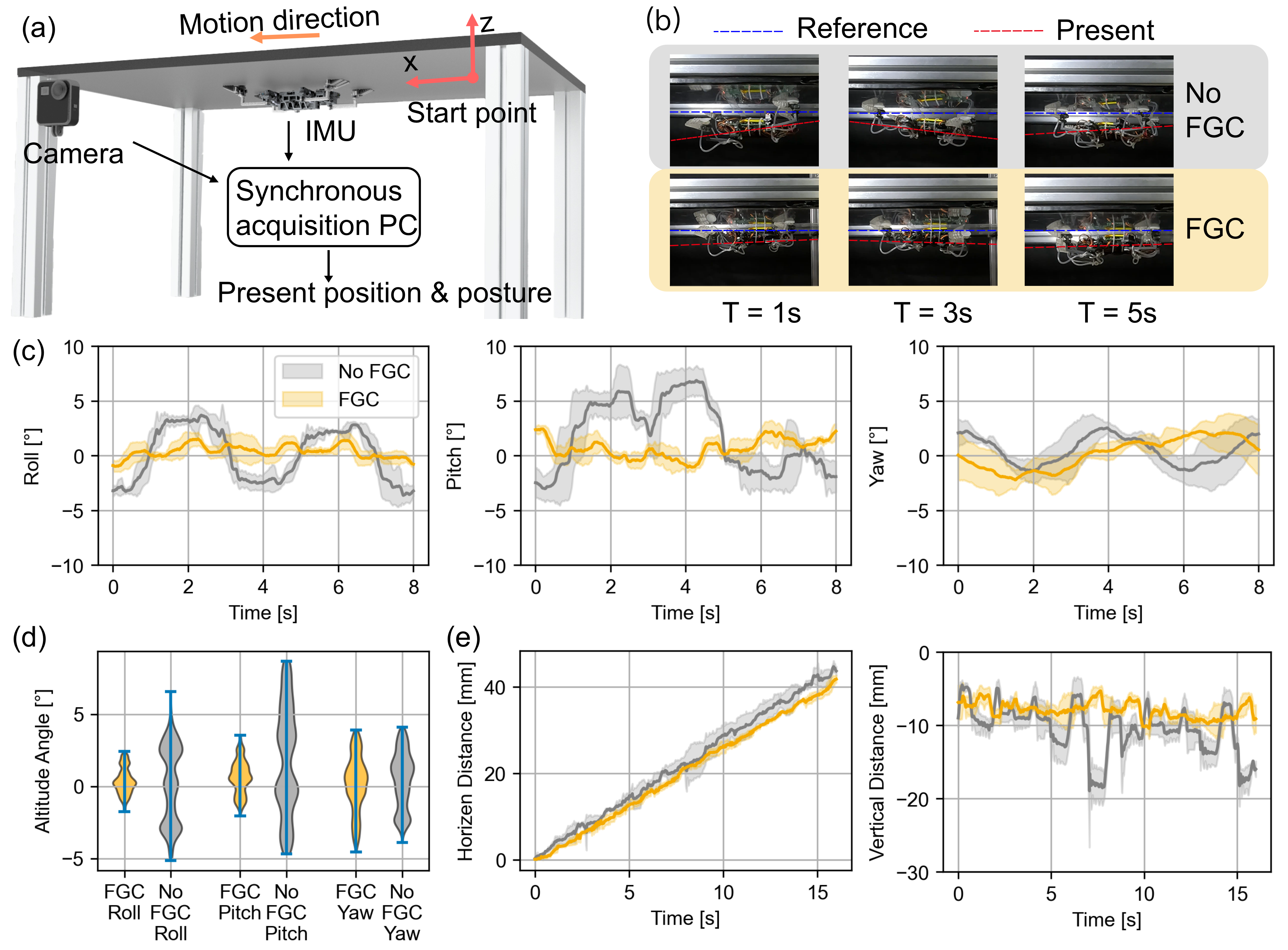}
    \caption{(a) The robot and 180-degree inverted platform. The hanging weight indicates that gravity is downward. (b) Screenshot of the motion capture. The blue dashed line represents the horizontal reference line, and the red dashed line represents the pitch angle of the robot. (c) Attitude angle. The gray solid line represents the theoretical value, while the black vertical solid line represents the boundary of the period. (d) The violin plot of attitude angle. The square in the box represents the mean value, and the horizontal line represents the median value. (e) The horizontal and vertical displacements of the robot over two gait circles.}
    \label{fig:platform}
\end{figure*}

In this section, the environmental parameter affecting the experiments is the gravitational force acting on the robot. The crawling surface is flat and made of PMMA with a surface roughness of $0.4\mu m$. The platform using in experiment is as shown in Fig.~\ref{fig:platform} (a).

In contrast to minimizing the sum of all end-effector forces, achieving force and torque equilibrium is of greater importance. Therefore, the optimization parameters chosen in equation (\ref{eq:5}) are continuously tested. The lowest sampling frequency was employed in the experiments, i.e., sampling occurred after each phase transition. Consequently, following offline computation of posture compensation for the ambling gait, we obtained eight segments of compensation parameters, including the command angles for the motors in the stance phase corresponding to the respective time segment.  

In the experiments, the Inertial Measurement Unit (IMU) is positioned at the center of the robot's back to collect data on the robot's roll angle, pitch angle, and yaw angle. During the experiments, a camera captured footage at 60 frames per second, positioned parallel to the robot's crawling direction. (The video can be found in the supplement video 1.)

Some frames from the recorded videos have been shown in Fig.~\ref{fig:platform} (b). The recorded video was processed using the deep learning-based tool named ‘DLTdv digitizing tool’ \citep{Hedrick2008,theriault2014protocol} to track the position of the robot's geometric center and foot. The data obtained was converted from pixel count to actual millimeter distances.

\subsection{Experimental procedures}

The experimental group adopts the FGC control strategy, which covers all aspects in Fig.~\ref{fig:diagram}. The control group named No FGC adopts the basic control method, which is the online computation without the compensation in Fig.~\ref{fig:diagram}. Except for different control methods, the gait, parameters, and experimental platform of the control group and the experimental group are the same, as described in the experimental preparation. To reduce disturbance and make the robot status of the control group and the experimental group the same, the same robot was used in the experiment, and the control strategy was switched between the control group and the experimental group through software download. For data recording, the start time of the robot's second cycle was used as the zero point of the time coordinate axis. The specific steps of the experiment can be found in the supplement document 3.

\section{Results}

We conducted 10 experiments on each of the two methods. The experimental group that did not use FGC only completed 3 experiments in full, with the rest failing during the operation, while the experimental group that used FGC successfully completed all the experiment tests. Fig.~\ref{fig:platform} (c) compares attitude angle data between the group without the FGC strategy and the group utilizing the FGC strategy.

% \begin{Fig}
%     \centering
%     \includegraphics[width=\linewidth]{fig/Fig6.jpg}
%     \caption{(A) Screenshot of the motion capture. The blue dashed line represents the horizontal reference line, and the red dashed line represents the pitch angle of the robot. (B) Attitude angle. The gray solid line represents the theoretical value, while the black vertical solid line represents the boundary of the period.}
%     \label{fig:res}
% \end{Fig}

When FGC strategy was not adopted, the roll angle reaches a mean of the maximum of 3.68 degrees and a minimum of -3.39 degrees, the average of the maximum pitch angle of the robot is 6.72 degrees, with a minimum of -3.10 degrees, while the yaw angle has a maximum of 2.42 degrees and a minimum of -1.31 degrees. When using the FGC strategy, the maximum roll angle is reduced to 1.41 degrees, with a minimum of -0.95 degrees. The pitch angle has a maximum of 2.11 degrees and a minimum of -1.14 degrees, while the yaw angle reaches a maximum of 2.14 degrees and a minimum of -2.11 degrees.

Comparing the use of the FGC strategy to not using it, the peak-to-peak values of roll and pitch angles are reduced by approximately 2.60 times and 3.18 times, respectively. It is visually apparent that using the FGC strategy results in smaller angles of pitch and roll, and the overall trend tends to be gradual. The data of attitude angles is presented using violin plots in Fig.~\ref{fig:platform} (d), clearly showing that the data using FGC is more concentrated around zero, while the data not using FGC is more evenly distributed.

The horizontal and vertical displacements of the robot are shown in the Fig.~\ref{fig:platform} (e). In terms of vertical displacement, robots with FGC always maintain a fluctuation range of 0.5 mm in the vertical displacement of the robot's body center, while those without FGC reach a fluctuation range of 2 mm. Despite significantly enhanced motion stability, the horizontal speed of the robot is almost unaffected, reaching a speed of 2.6 mm/s in an ambling gait.

\section{Discussions}

Based on the experimental results, which includes peak-to-peak values of body posture angles, and the distance from the geometric center of the robot's body to the climbing plane, using the FGC strategy clearly makes the robot's body stable to the climbing plane. It means a more stable body posture compared to not employing the strategy. This validates the effectiveness of the FGC strategy.

The FGC strategy is not limited to quadruped climbing robots; its flexibility can extend to other robot types, including those with three or more leg configuration. This adaptability arises from the leg stiffness model within the FGC strategy, which can be represented using the DH method. The DH method can describe all series configurations. Conversely, the FGC strategy can address the challenge of unreliable attachment of the end-effector to the contact surface due to leg stiffness without additional structures and components. This approach is particularly suitable for robots with specific angular requirements between the end-effector and the contact surface.

\begin{table*}
 \caption{Wall-climbing robots parameters and performances table. BL represents body length, NG represents not give, and $^*$ represents estimated data from the figure of corresponding source. 'Independent' means the all components like controller and power supply are embedded in the robot. The maximum speed of our robot is measured with supplement video 2.}
  \begin{tabular}[htbp]{@{}ccccccccc@{}}
    \hline
    Robot & \makecell{BL\\(cm)} & \makecell{Mass\\(g)} & Attachment & \makecell{Climbing \\angle ($^\circ$)} & \makecell{Max speed\\(BL/s)} & Control & \makecell{Max pitch\\($^\circ$)} & Characteristic \\
    \hline
    Ours & 18 & 487 & \makecell{Adhesion-\\suction} & 0-180 & 0.051 & Open & 2.11 & \makecell{independent,\\ stable}\\
    \cite{Hawkes2011} & 60 & 370 & \makecell{Adhesion} & 0-90 & 0.067 & Open & NG & \makecell{independent,\\ stable}\\
    \cite{ko2017wall}  & 23 & 363 & \makecell{Adhesion} & 0-180 & 0.0043 & Open & NG & \makecell{independent, \\ waterproof}\\
   
    \cite{chen2020inverted} & 4.5 & 1.45 & \makecell{Electro-\\adhesion} & 0-180 & 0.01 & Open & \makecell{3.03$^*$} & \makecell{light}\\
     \cite{Hong2022} & 33 & 8000 & \makecell{Electro-\\magnetic} & 0-180 & 1.51 & Closed & \makecell{2.86$^*$} & \makecell{independent,\\ wall-to-wall\\transition}\\
    \cite{10252004} & 27.5 & 1190 & \makecell{Adhesion} & 0-180 & NG & Open & \makecell{2.40$^*$} & \makecell{high load\\ capacity}\\
    % Row 1, Col 1  & Row 1, Col 2  & Row 1, Col 3  \\
    % Row 2, Col 1  & Row 2, Col 2  & Row 2, Col 3  \\
    \hline
  \end{tabular}
\end{table*}

Table 1 shows a comparison of the proposed robot and other climbing robots based on the robot parameters and key performance. The development of robots presents challenges due to constraints on mass, size, and power consumption, leading to limitations in structural design, computational power for control, and choice of drive modes. Small climbing robots all adopt the open loop gait towing to the weight, size, and power consumption limitations of miniaturized robots. This article studies the ceiling climbing of robots under such limited conditions and proposes the FGC strategy to enhance climbing stability. MARVEL and MST-Q are medium-sized robots, while the rest are small robots and independence is more difficult to achieve for small robots. Compared to externally assisted robots, although MST-Q has higher load capacity and HAMR-E has ideal speed, it does not need to bear too much self load and has powerful external energy sources. Although the high voltage required for the electro adhesion material used in HAMR-E has been reduced to 250V, a tether is still needed for external power supply. MST-Q can crawl on an inverted surface and has strong load-bearing capacity, but from its video, it seems that the speed is not so satisfactory. Compared to the medium-sized robot MARVEL, its speed is faster due to having more space for structural adjustment, improving its motion efficiency, and its magnetic attachment method determines the speed of attachment and detachment, but also limits the crawling surface to be magnetic. Our robot can independently climbing on inverted surfaces, not limited to magnetic surfaces, and has a small mass and size.

The pitch angle is used to compare the stability of the robot's climbing performance. Frame by frame analysis was conducted on all robots in Table 1, the maximum pitch angle of the robot can be estimated. And the maximum pitch angle of our robot is obtained through IMU. It can be seen that the robot EF-\uppercase\expandafter{\romannumeral1} using FGC strategy has the smallest maximum pitch. It indicates that the body of our robot EF-\uppercase\expandafter{\romannumeral1} is more stable during climbing.

The biggest limitation of FGC at present is that it relies on offline computation. Since calculating the leg compensation requires extensive computation, it has to rely on the results of prior offline computations. This makes it difficult to obtain updated compensation after online adjustments to gait and foot trajectory.

\section{Conclusion}
In this study, a leg-enhanced stiffness model was established by analyzing the leg deformation, and a feedforward gravity compensation limb body coordination strategy was proposed. This strategy used quadratic programming to optimize the foot force based on the state of the legs in the stance phase, achieving coordination of the supporting leg. Finally, the forward feedback method was used to correct the body tilt, making it easier for the swinging leg to achieve a reliable attachment to the target plane. Owing to its offline computing characteristics, this strategy could be applied to small robots with limited computing power. This study conducted experiments using a quadruped climbing robot on an inverted plane, employing the ambling gait. This strategy not only made robot climbing more stable, but also improved the success rate of robot crawling on inverted surfaces.

Due to the high computational demands of FGC optimization and matrix calculations, the algorithm still cannot run in real-time on robots and relies on offline computation. Therefore, we plan to use neural networks such as long-short term memory in our future work to perform regression analysis on previously computed results, thereby accelerating the compensation process to achieve real-time performance.

%%%%%%%%%%%%%%

% \begin{appendices}

% \end{appendices}

\section{Competing interests}
No competing interest is declared.

 \section{Acknowledgments}

This research was supported by National Key R\&D program of China (2023YFE0207000), the National Natural Science Foundation of China (Grant No. 51975283 and 62233008) and a Swiss NSF grant
to A.J.

 \section{Data availability statement}

 The data and code that support the findings of this study are openly available at the following URL: https://github.com/bishopAL/FeedforwardGravity\\Compensation-WCR.

\bibliographystyle{agsm}
%\bibliographystyle{unsrt}
% \bibliography{2023_BB,reference}
\bibliography{FGC}

\end{document}